\documentclass[letterpaper, 10 pt, conference]{ieeeconf}  % Comment this line out if you need a4paper

\IEEEoverridecommandlockouts                              % This command is only needed if 
\usepackage{graphics} % for pdf, bitmapped graphics files
\usepackage{epsfig} % for postscript graphics files
\usepackage{mathptmx} % assumes new font selection scheme installed
\usepackage{tabularx}
\usepackage{times} % assumes new font selection scheme installed
\usepackage{amsmath} % assumes amsmath package installed
\usepackage{amssymb}  % assumes amsmath package installed
\usepackage{hyperref}

\usepackage[]{xcolor}
\title{\LARGE \bf
A Learning-Free Characterization Framework for the Resilience and Sensitivity of Polyurethane Vision-Based Tactile Sensors
}

\author{Benjamin Davis and Hannah Stuart*% <-this % stops a space
\thanks{This material is based upon work supported by Google and the National Science Foundation Graduate Research Fellowship Program under Grant No. 2146752. Any opinions, findings, and conclusions or recommendations expressed in this material are those of the author(s) and do not necessarily reflect the views of the National Science Foundation. }
\thanks{Benjamin Davis and Hannah Stuart are with the Embodied Dexterity Group, Dept. of Mechanical Engineering, University of California Berkeley, Berkeley, CA, USA (e-mails: bendavis, hstuart@berkeley.edu).}
}

\begin{document}

\maketitle
\thispagestyle{empty}
\pagestyle{empty}

%%%%%%%%%%%%%%%%%%%%%%%%%%%%%%%%%%%%%%%%%%%%%%%%%%%%%%%%%%%%%%%%%%%%%%%%%%%%%%%%
\begin{abstract}

Vision-based tactile sensors (VBTSs) are a promising technology for robots, providing them with dense signals that can be translated into a multi-faceted understanding of contact. However, existing VBTS tactile surfaces make use of silicone gels, which provide high sensitivity but easily deteriorate from loading and surface wear. Furthermore, existing literature lacks rigorous durability and sensitivity evaluations targeted for intrinsic sensor performance. We propose that polyurethane rubber, a typically harder material used for high-load applications like shoe soles, rubber wheels, and industrial gaskets, may provide improved physical gel resilience, potentially at the cost of sensitivity. In addition, we propose a methodological framework to evaluate and compare tactile sensor hardware across designs and material compositions. Our resilience tests assess sensor durability across normal loading, shear loading, and abrasion. For sensitivity, we introduce learning-free assessments of force and spatial sensitivity to isolate intrinsic sensor capabilities from the confounding effects of downstream dataset and network architecture choices. We also perform a system-level validation using a bottle cap loosening and tightening task to show the translation of our controlled test results with a real-world example. Our results show that polyurethane substantially improves resilience. While it sacrifices sensitivity at low forces, the effective force range is largely increased, revealing the utility of polyurethane VBTSs over silicone versions in more rugged, high-load applications.

\end{abstract}

%%%%%%%%%%%%%%%%%%%%%%%%%%%%%%%%%%%%%%%%%%%%%%%%%%%%%%%%%%%%%%%%%%%%%%%%%%%%%%%%
\section{Introduction}

The importance of tactile sensing in robotics is becoming increasingly recognized as robots transition from controlled laboratory environments to more unstructured and unpredictable scenarios like homes, factories, or elsewhere. Tactile sensing, the robotic equivalent of the human sense of touch, provides information about physical interactions with objects and the environment that allows robots to perform complex manipulation tasks, handle delicate objects, adapt to their surroundings, and recover from mistakes. Extensive research has been performed on the development and improvement of a range of tactile sensing technologies, including capacitive/piezoresistive/piezoelectric sensors, barometric sensors, and vision-based (or optical) tactile sensors \cite{kappassov_tactile_2015,Cutkosky2016}.

\subsection{Vision Based Tactile Sensors}

Among these technologies, vision-based tactile sensors (VBTSs) are increasingly promising due to the rich information they provide (e.g., Digit \cite{lambeta_digit_2020}, Gelsight \cite{yuan_gelsight_2017}). These sensors operate by using an internal camera to observe the deformation of a soft elastomer that physically contacts the environment. The gel's deformation is captured by an image transfer layer--often reflective and/or covered with a pattern of markers--which is cast onto a transparent base layer. This assembly is adhered to an acrylic window in front of the internal camera \cite{zhang_hardware_2022}. In existing work, both layers are typically made with silicone, a soft elastomer that provides high sensitivity during contact. VBTSs offer multiple advantages over more traditional tactile sensors, including high-resolution data, reduction of custom wires and circuitry, and the removal of electronics from the tactile interface itself \cite{zhang_hardware_2022}. The visual data from the sensor can be processed to extract many types of information, such as contact region, forces, texture classification, pose estimation, and slip detection \cite{zhang_deltact_2022, baghaei_naeini_novel_2020, andrussow_minsight_2023, lin_dtact_2022, kakani_vision-based_2021, zhang_fingervision_2018}. 

\begin{figure}[t]
    \centering
    \vspace{3mm}
    \includegraphics[width=85mm]{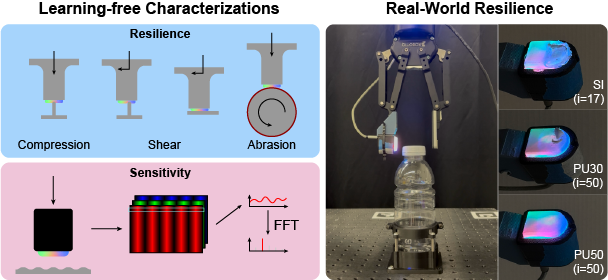}
    \caption{In this work, we propose a series of learning-free characterizations across inherent sensor resilience and sensitivity and demonstrate translation of their results to a real-world application.}
    \label{fig:frontpage}
\end{figure}

\subsection{Challenges in Durability and Characterization}

Despite their advantages, a major limitation of VBTSs is the durability of their silicone gels. These gels have multiple known failure modes, including surface wear of the image transfer layer via abrasion/scratching/tearing and delamination of the silicone from the acrylic window \cite{zhao_polytouch_2025, wang_gelsight_2021, potdar_high-speed_2024}. While their performance may be suitable for in-lab data collections and demonstrations, their deployment in the real world requires sensors that can endure repeated and unexpected use. Prior work has sought to improve gel durability with protective coverings like latex or tape \cite{wang_gelsight_2021, potdar_high-speed_2024, rayamane_design_2022, yamaguchi_combining_2016}, though these often induce image artifacts \cite{wang_gelsight_2021, potdar_high-speed_2024}. Another approach is to select higher hardness silicones \cite{zhang_hardware_2022}, but this does not address the issue of delamination resulting from low bond strength of silicone to acrylic. As a result, most existing VBTS applications are limited to low-force manipulation tasks. In practice, these sensors may endure higher loads--possibly unexpectedly--and require an increased level of sensor resilience to various forms of loading and abrasion. 

Beyond physical limitations, the field also lacks sufficient standardized characterization frameworks to systematically compare durability and sensitivity across VBTS hardware. Methodological durability assessments are rare in literature and focus only on a single, specific mode of failure \cite{lambeta_digit_2020, rayamane_design_2022, zhao_polytouch_2025}, rather than testing across a spectrum of wear modes expected during real-world deployment. Additionally, while recent years have yielded many evaluations of VBTS performance \cite{liu_tip_2025, higuera_sparsh_2024, schneider_tactile_2025}, they are overwhelmingly learning-based. As a result, sensor performance depends heavily on model selection, datasets, and selected robotic tasks, which can obscure the sensor hardware's intrinsic physical capabilities. While these learning-based evaluations are useful for assessing performance in applied contexts, we argue that a complementary, learning-free evaluation can enable quicker, simpler baseline comparisons of sensitivity across different hardware platforms.

\subsection{Proposed Materials and Methodological Framework}

To address these two challenges, we first propose a novel polyurethane gel as a highly resilient alternative to silicone for VBTSs. Polyurethane, an elastomer used in various high-wear applications like shoe soles, skateboard wheels, and industrial gaskets, offers a promising alternative to silicone for improving VBTS durability. Subject to the requirement of transparency (for the base layer), readily available polyurethane rubbers have higher hardnesses than commonly used clear silicones, providing an intuitive advantage against mechanical wear. Importantly, polyurethane also bonds well to acrylic via superglue, mitigating the delamination failures common to silicone. Despite these apparent advantages, polyurethane VBTS gels remain largely unexplored in current literature.

To systematically evaluate our proposed polyurethane gels against a standard silicone baseline, we introduce a reproducible, learning-free methodological framework testing across mechanical resilience and signal sensitivity. For resilience, we propose a collection of mechanical tests evaluating repetitive compressive loading, shear loading, and abrasion to simulate a lifetime of wear. For sensitivity, we introduce learning-free evaluations of force and spatial sensitivity to allow for baseline comparisons across silicone and polyurethane VBTS gels. We use frequency domain analysis to extract signal strength from a ridged surface, taking inspiration from human grating orientation discrimination tasks \cite{johnson_tactile_1981, van_boven_limit_1994, craig_grating_1999}. A recent approach similarly performs a grating test for tactile sensor evaluation but relies on the Structural Similarity Index Measure (SSIM) and support vector machine (SVM) classification \cite{liu_tip_2025}. However, this approach still relies on data collection and model training, and the use of SSIM (which reflects human visual perception) may present biases compared to the use of raw sensor data.

\subsection{Overview}

We present two primary contributions in this work:
\begin{enumerate}
    \item The formal introduction and accessible fabrication methodology of polyurethane VBTS gels for improved sensor resilience in high-wear applications.
    \item A reproducible, learning-free sensor methodological framework for quantifying intrinsic VBTS capabilities, encompassing physical resilience (across compressive, shear, and abrasive wear) and force/spatial sensitivity.
\end{enumerate}
We begin by describing both our baseline silicone and polyurethane gel fabrication techniques in Section \ref{sec:gel_fab}. Then, we present the parameters and procedures for our resilience and sensitivity characterizations in Section \ref{sec:evaluation}. Results describing the tradeoff between resilience and sensitivity are detailed in Section \ref{sec:results}, and outcomes are discussed in Section \ref{sec:discussion}. Finally, we present a system-level validation with a bottle cap unscrewing task in Section \ref{sec:demo} to ensure our intrinsic characterizations translate with real-world use before concluding in Section \ref{sec:conclusion}.

\begin{figure*}[t]
    \centering
    \vspace{3mm}
    \includegraphics[width=165mm]{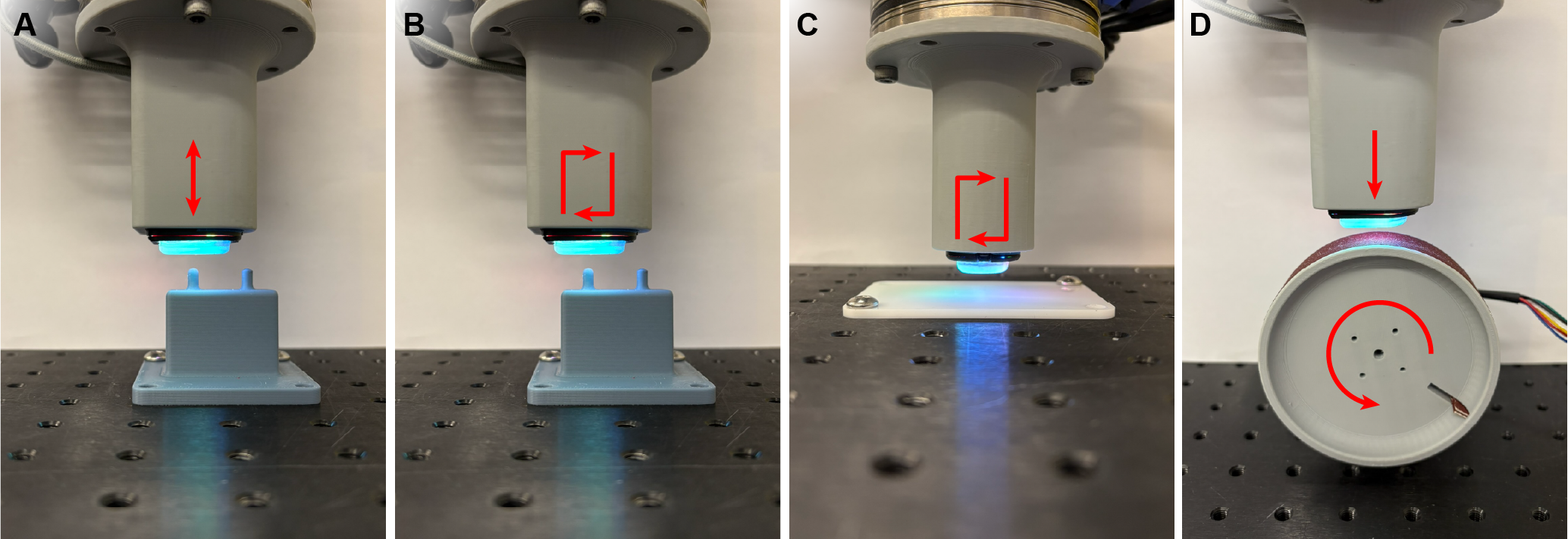}
    \caption{We perform four resilience tests: (A) cyclic compression on an indenter, (B) cyclic local shear on an indenter, (C) cyclic transverse shear on a flat surface, and (D) abrasion. The setup in (A) is also used to characterize the sensor's force sensitivity.}
    \label{fig:resilience_tests}
\end{figure*}

\section{Gel Fabrication}
\label{sec:gel_fab}

In this section, we describe our fabrication techniques for producing the silicone and polyurethane gels used for comparison. We use the body of the Digit \cite{lambeta_digit_2020} as a base, which contains a camera and three RGB LEDs. It uses a gel composed of a transparent silicone base layer (Smooth-On Solaris, Shore A 15) and an opaque image transfer layer (Smooth-On EcoFlex 00-10, Shore 00-10) colored with white pigment and airbrushed onto the base layer. This gel is glued onto an acrylic window with Smooth-On Sil-Poxy. The acrylic-gel unit can then be press-fit into the Digit case. Due to the Digit's modularity and open-sourced design, we are able to develop and test custom acrylic-gel units with silicone and polyurethane. We make our gels flat rather than rounded like the original Digit gels for ease of fabrication and better compatibility with our tests. We describe our manufacturing processes for each material below.

\subsubsection{Silicone Gels}

For a silicone baseline, we replicate the material composition of the Digit gel due to its prevalence and similarity in hardness to other commonly used VBTSs like the GelSight family \cite{zhang_hardware_2022}. We first create the silicone base layer with equal parts A and B (by weight) of Smooth-On Solaris. The mixture is degassed and then poured into a 4mm deep 3D printed mold to cure. To form the image transfer layer, we combine equal parts A and B of EcoFlex 00-10. We then mix in 10\% by weight of Smooth-On Silicone Thinner, which reduces layer thickness for improved sensitivity, and 1\% by weight of Smooth-On Cast Magic Silver Bullet powder and 2\% by weight of Smooth-On Silc Pig white pigment, both of which contribute to lower opacity and higher reflectivity. To combine the base layer and image transfer layer, we place the cured base layer facing upwards on a flat acrylic plate, and the uncured image transfer layer mixture is poured over until it is covered. After this step, we leave the silicone to cure for four hours. Once completely cured, we use a razor blade to clean up the edges of the gel before gluing it to a 6.35mm thick clear acrylic window with Smooth-On Sil-Poxy.

\subsubsection{Polyurethane Gels}

We fabricate two versions of a polyurethane gel using the same technique. The harder version uses Smooth-On Clear Flex 50 (Shore A 50) for the base layer and Smooth-On Vytaflex 40 (Shore A 40) for the reflective layer, while the softer version uses Smooth-On Clear Flex 30 (Shore A 30) and Smooth-On Vytaflex 20 (Shore A 20) in their place. We select these formulations since Clear Flex 30/50 are the softest readily available water-clear polyurethanes. To form the base layer of the polyurethane gel, we use the specified ratio of Clear Flex Parts A and B, mix thoroughly, and degas before curing in a silicone negative mold. The image transfer layer is created by mixing equal parts A and B of VytaFlex. To this mixture, we add 1\% by weight of Smooth-On Cast Magic Silver Bullet powder and 2\% by weight of Smooth-On SO-Strong white urethane colorant, mix thoroughly, and degas under vacuum. To attach the base and image transfer layers, we first pour 0.25 g of the uncured image transfer layer mixture into the base of the emptied silicone mold. The specific volume of VytaFlex 20/40 is $1000.7/971.8~\frac{\text{mm}^3}{\text{g}}$, yielding a layer height of approximately 0.64/0.62 mm (not considering mix-ins or mold fillets). We then use an air blow gun to gently spread the mixture along the base of the mold and remove large bubbles by popping with tweezers. Next, we place the base layer back into the mold and press it onto the uncured image transfer layer, ensuring that no air is captured in pockets between the layers. Once placed, the gel is left to cure. The cured gel is then attached to an acrylic window by spreading superglue across the surface of a window, pressing the gel on firmly, and wiping away excess glue.

\begin{figure*}
    \centering
    \vspace{3mm}
    \includegraphics[width=175mm]{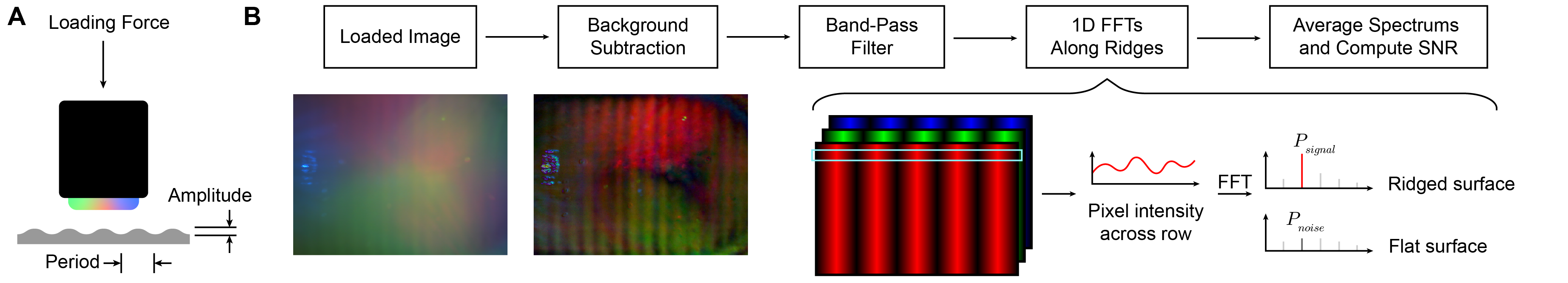}
    \caption{(A) Our learning-free spatial sensitivity evaluation uses the sensor reading when pressed onto a ridged surface of varying period and amplitude. (B) The image is preprocessed via background subtraction and bandpass filtering before being cropped and run through a series of 1D FFTs. The resulting power spectral densities are averaged. The signal power is compared to a noise power, defined by the same frequency range from a flat, ridgeless surface, to obtain a signal-to-noise ratio (SNR). The background subtraction image is amplified for visualization purposes.}
    \label{fig:img_processing}
\end{figure*}

\section{Sensor Characterization Framework}
\label{sec:evaluation}

We present a series of reproducible tests to characterize the resilience and sensitivity of each gel. The hardware necessary for each test is 3D-printable, and part files are available at \href{https://github.com/benudavis/resilient_vbts_hardware}{github.com/benudavis/resilient\_vbts\_hardware}.

\subsection{Mechanical Resilience Characterization}

To characterize the mechanical resilience of our silicone and polyurethane gels, we put them through a series of mechanical tests representative of common wear modalities that may be encountered during use, including compression, shear, and abrasion (Fig. \ref{fig:resilience_tests}). For these tests, we select loading forces that 1) provide a good spread of data, and 2) are at or near those of human grasping and manipulation, a popular application for such sensors. In future work, reparameterization of these tests to represent loading conditions in different applications (e.g. medical or manufacturing) should be explored. We perform all tests with the sensor mounted to a Universal Robots UR-10 robot arm.

\subsubsection{Cyclic Compression Loading}

We first test the gel's ability to withstand stretching or puncture from repeated compressive loading with the setup in Fig. \ref{fig:resilience_tests}A. Compressive loads may occur often during the lifetime of the sensor as it is used to grasp objects. We perform this test on a 4 mm spherical tip indenter with a 15 N compressive load and 1000 cycles. Testing with the narrow indenter, similar to grasping objects with sharper points or edges, provides a conservative estimate of resilience to compressive wear. The 15 N load is on the high end of typical human grasp forces \cite{rajakumar_evidence_2022} but was the minimum load where failure of any sensor was observed. We test for 1000 cycles to reasonably approximate a lifetime of sensor use without requiring an extensive amount of time.

\subsubsection{Cyclic Shear Loading}

We also test the gel's resilience under shear. We include both local shear and transverse shear, or shear across the whole face of the sensor. For local shear, we compress the sensor onto a 4 mm spherical tip indenter with a 10 N compressive load, and then apply a 5 N lateral load. This setup is shown in Fig. \ref{fig:resilience_tests}B. These loads are again representative of human grasp forces \cite{rajakumar_evidence_2022} and help to isolate the effect of local shear, as pilot testing revealed that all materials could withstand 10 N of cyclic compression alone. Local shear loads could occur when a grasped object with small contact points is subjected to unexpected external loads and may induce failures like tearing or puncture. For transverse shear, we compress the sensor onto a flat acrylic plate with a 15 N compressive load and then apply a 15 N lateral load, as seen in Fig. \ref{fig:resilience_tests}C. We apply a 15 N compressive load to increase friction forces and prevent the gel from slipping on the plate. This form of loading can cause delamination of sensor layers and is also important to investigate. The tests for local shear and shear across the whole gel are both run for 1000 cycles.

\subsubsection{Abrasion}

To test the gel's abrasion resistance, we abrade it with a custom setup shown in Fig. \ref{fig:resilience_tests}D. We wrap 150 grit sandpaper around a 3D printed 95.5 mm diameter wheel controlled with a brushed DC motor. The robot arm presses the gel against the sandpaper with 5 N of force. We then spin the wheel against the gel at constant velocity for 8 m in 2 m increments, where an unloaded sensor image is captured after each increment. The sandpaper is replaced before every full 8 m trial and used for its entire duration. While this form of abrasion may be extreme compared to actual applications, it provides a  standard way to compare sensor resilience for a worst-case representation of wear.

\subsubsection{Evaluation}

We evaluate all resilience tests with the same metric. After each cycle, we record 10 RGB frames of the unloaded sensor reading at 30 frames per second with QVGA resolution (320x240 pixels) and average them to get one image per cycle. To evaluate sensor damage, we calculate the mean absolute error ($MAE$) between these images at the current cycle and the first cycle:
\begin{align}
    MAE_{n} = \frac{1}{320 \times 240 \times 3}\sum_{i=1}^{320} \sum_{j=1}^{240} \sum_{c=1}^3 |I_n(i,j,c) - I_1(i,j,c)|
\end{align}
where $n$ represents the current cycle, $i$ and $j$ are pixel coordinates, $c$ is the color channel, and $I_n$ and $I_1$ are the current cycle and first cycle images, respectively. We select this metric because it directly measures changes in the sensor reading, revealing deterioration in sensor transduction over time. 

\subsection{Sensitivity}

Improving sensor resilience typically reduces sensitivity. To assess this in our polyurethane gels, we propose two tests to evaluate force sensitivity and spatial sensitivity.

\subsubsection{Force Sensitivity}

We evaluate the force sensitivity of our different sensor gels by loading them up to 40 N on a 4 mm spherical tip indenter at a rate of 2e$^{-6}$ m/s. We use the same setup used for cyclic compression (Fig. \ref{fig:resilience_tests}A). We restrict the maximum load to 40 N due to hardware limitations, although this range is sufficient for distinguishing sensor performance. During a single loading and unloading, we record 320x240 RGB frames at 30 frames per second with the Digit sensor and normal forces with a force torque sensor (ATI Axia80-M8). We then calculate the $MAE$ of each frame with respect to the first frame to quantify how much the captured image changes with loading.

\subsubsection{Spatial Sensitivity}

Our approach for assessing spatial sensitivity is a novel, learning-free method that analyzes features directly rather than relying on model training. We accomplish this by performing a frequency-domain analysis of the sensor's measurement of a periodic, ridged surface. By varying the period of the ridges, we can assess x/y-sensitivity; similarly, we can evaluate z-sensitivity by varying amplitude. The process is depicted in Fig. \ref{fig:img_processing}.

First, the sensor is pressed against the ridged surface with a known period and amplitude. We record 500 frames of RGB image data at 30 frames per second at a resolution of 640x480 pixels. We record another 500 frames with the sensor unloaded. The sequences of 500 frames are averaged to mitigate the effects of random noise, yielding one unloaded image and one loaded image.

We then preprocess the images to clean the signal. First, we perform background subtraction between the unloaded and loaded frames, $\Delta I = I_{loaded} - I_{unloaded}$, to isolate the changes observed due to sensor gel deformation. Next, the image is spatially filtered to remove low-frequency artifacts and high-frequency noise using a difference of Gaussians method. This involves defining two Gaussian kernels: a narrow kernel to smooth high-frequency noise, and a wide kernel to isolate low-frequency artifacts like uneven lighting:
\begin{align}
    G(x,y,\sigma) = \frac{1}{2 \pi \sigma^2} e^{-\frac{x^2+y^2}{2\sigma^2}}
    \label{eq:gauss_kernel}
\end{align}
where the standard deviation $\sigma$ determines the smoothing extent. To obtain the final filtered image $D(x,y)$, we convolve the delta image $\Delta I(x,y)$ with the difference of the two Gaussian kernels:
\begin{align}
    D(x,y) = \Delta I(x,y) * \left( G(x,y,\sigma_{low}) - G(x,y,\sigma_{high}) \right).
\end{align}
Finally, the image is cropped to limit effects caused by camera distortion and the edges of the gel.

To quantify the sensor's spatial response, we perform a series of one-dimensional Fast Fourier Transforms across each row of pixels spanning the ridged pattern for each color channel, centering to eliminate the DC component. We then calculate power spectrums and average across all rows and channels to obtain a single non-normalized power spectral density (PSD) for the sensor's signal. We use this spectrum to calculate spatial sensitivity using a Signal-to-Noise Ratio (SNR). The noise floor is obtained by recording the PSD on a flat, ridgeless surface. Next, we calculate signal power $P_{signal}$ for a ridged surface as the summed power of the two frequency bins closest to the known ridge frequency. We define the noise power $P_{noise}$ as the summed power of the corresponding frequency bins from the spectrum of the flat surface (Fig. \ref{fig:img_processing}). Finally, we calculate SNR in decibels (dB) as follows:
\begin{equation}
    SNR_{dB} = 10 log_{10} \left( \frac{P_{signal}}{P_{noise}} \right).
\end{equation}

In this study, we test across 10 ridge amplitudes with constant period, and 10 ridge periods with constant amplitude. All ridges were printed with custom settings on a Bambu X1E. We test across amplitudes of 0.005 to 0.05 mm, and across periods of 0.6 mm to 1.5 mm. Although absolute positioning variance may exist at this scale, we find that local path-following accuracy is sufficient to produce consistent ridges and elect to use it for its wide accessibility. We also use the exact same part across the three gels, ensuring that relative differences result from sensor performance and not differences in target geometry. Ridges of varying magnitude all have the maximum period of 1.5 mm, while ridges of varying period have the same maximum amplitude of 0.05 mm. We selected these ranges experimentally to capture transitions from sub-resolution features to clear ones, ensuring that all materials could generate a meaningful signal at the largest period and amplitude. The normal force applied between the sensor and surface is also important for the strength of the signal. Because harder materials will deform less and thus produce a weaker signal compared to softer materials, we also test both 2 N and 10 N normal forces across all ridge configurations.

\section{Results}
\label{sec:results}

\begin{figure*}
    \centering
    \vspace{3mm}
    \includegraphics[width=175mm]{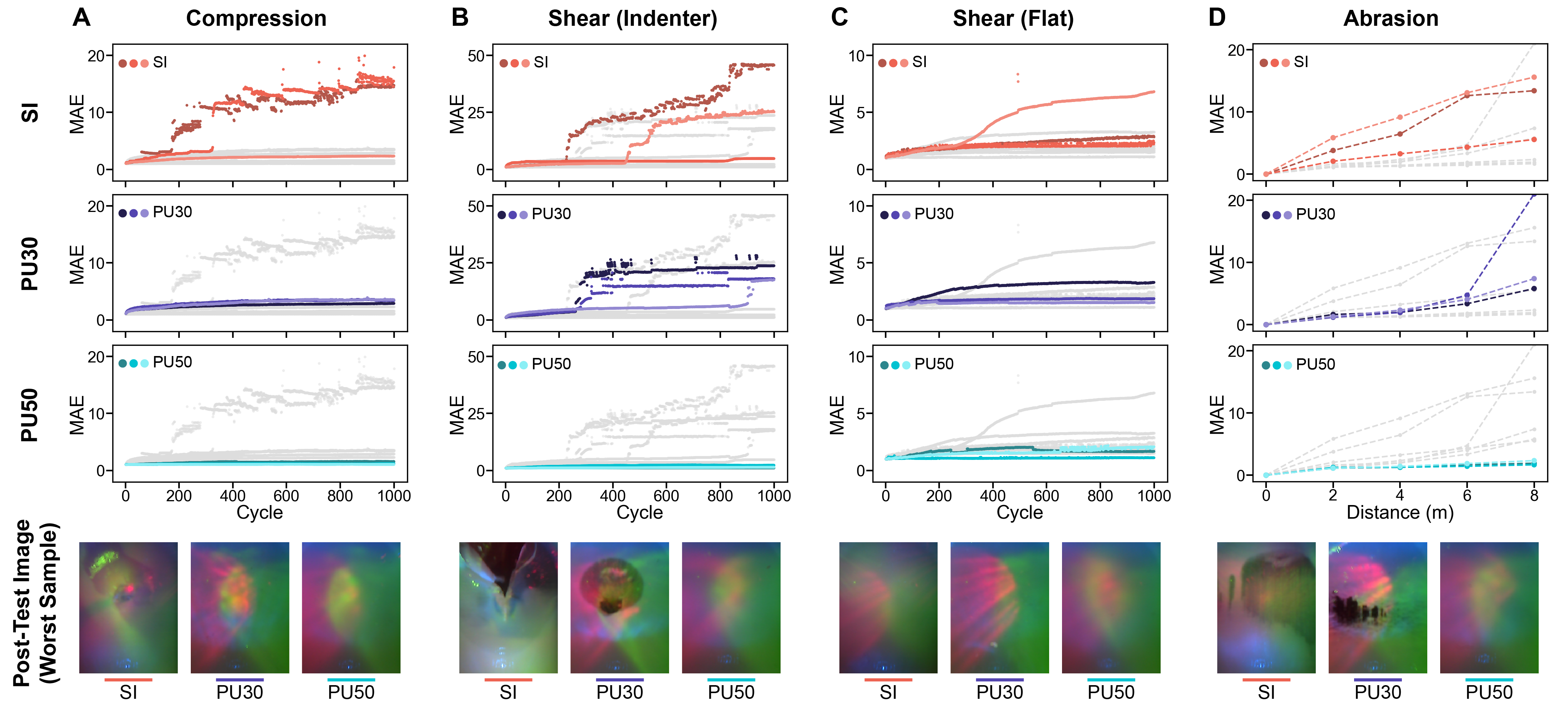}
    \caption{Results for cyclic compression (A), cyclic shear on an indenter (B), cyclic shear on a flat surface (C), and abrasion (D) tests across three gel samples for silicone (SI; shown in the top row in red), a 30A hardness polyurethane (PU30; shown in the second row in purple), and a 50A hardness polyurethane (PU50; shown in the third row in cyan). Each color shade represents a different sample, and grey dots represent results from the other materials. The raw sensor images in the fourth row depict the final sensor image (after 1000 cycles) from the worst-performing sample of each material. Different sets of gels are used for each test.}
    \label{fig:test_results}
\end{figure*}

\subsection{Mechanical Resilience}

Resilience testing results (Fig. \ref{fig:test_results}) reflect evaluations across three independent samples per material per condition, yielding 36 unique samples total (not counting those used for pilot testing). While our sample size prevents useful statistical testing, it demonstrates macro-level variance and primary failure modes. We include representative post-test sensor images from the worst-case sample of each material.

\subsubsection{Cyclic Compression Loading} 

Under cyclic compression (Fig. \ref{fig:test_results}A), two of three silicone samples fail catastrophically due to puncture within 250-500 cycles, leading to large increases in $MAE$. The third sample undergoes a much smaller increase in MAE throughout the test. This gel's image transfer layer undergoes permanent visible deformation from the indenter, but puncture does not occur. The three 30A hardness polyurethane (PU30) gels see minor change across cycles but perform much better than the SI gels on average. The 50A hardness polyurethane (PU50) gels remain largely stable during the test, consistently outperforming both other materials.

\subsubsection{Cyclic Shear Loading} 

Under local shear (Fig. \ref{fig:test_results}B), two silicone samples encounter catastrophic failure with puncture and tearing occurring before 500 cycles of loading. The third sample is also punctured, but the hole is clean and only induces a small increase in $MAE$ relative to the other two silicone samples. The PU30 gels all experience some degree of failure via tearing of the reflective image transfer layer, showing little to no performance benefit over SI. Again, the PU50 gels consistently outperform the other two materials, seeing minimal changes in reading throughout the test. 

For transverse shear on a flat surface, results in Fig. \ref{fig:test_results}C show limited change in reading across the 1000 cycles. The SI gels undergo varying degrees of minor delamination from the acrylic window, with one sample seeing a particularly large amount. Still, the $MAE$ is limited to $\sim$7 units per pixel, which is much less than that observed for cyclic compression and shear failures with the indenter. PU30 slightly outperforms SI on average but still shows some change over time; however, no visible delamination was observed upon inspection. The PU50 gels provide relatively constant images throughout the cycles, with no visible delamination present.

\subsubsection{Abrasion}

Abrasion testing results are shown in Fig. \ref{fig:test_results}D. The SI gels experience varying degrees of tearing in the image transfer layer, often occurring within 2 m of abrasion and increasing with more distance. PU30 largely shows better resistance to abrasion than SI, although one sample undergoes tearing in the final 2 m and sees a large increase in MAE. PU50 outperforms SI and PU30, showing little change in MAE for all three samples. However, the gels still exhibited obvious signs of wear. During testing, the polyurethane image transfer layer wore away as small particles rather than tear in bulk like their silicone counterparts, leading to this improvement in performance. 

\subsection{Sensitivity}

\begin{figure}
    \centering
    \vspace{3mm}
    \includegraphics[width=85mm]{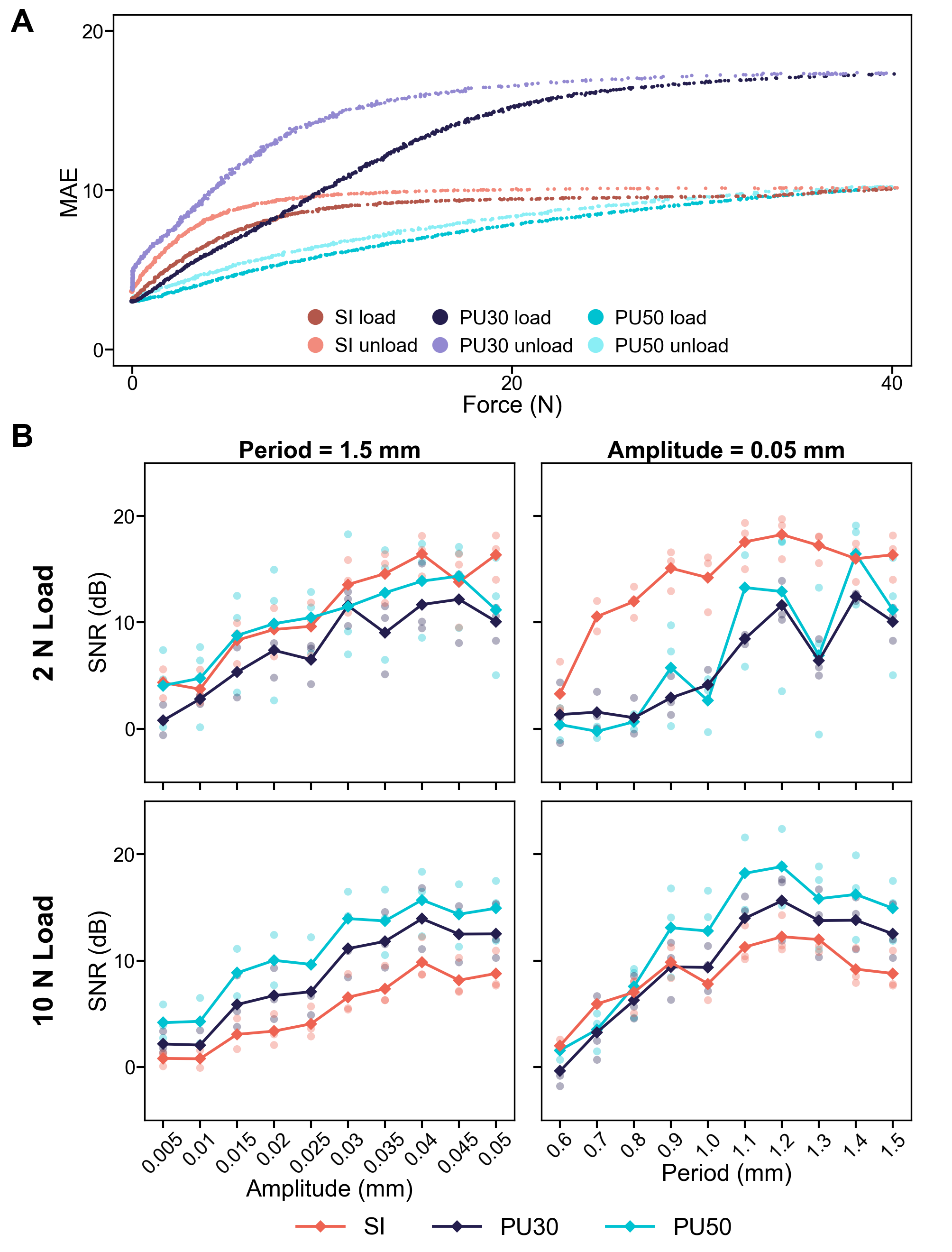}
    \caption{Results for force sensitivity (A) and spatial sensitivity (B) tests. In (A), the mean absolute error (MAE) is calculated with respect to the unloaded image to quantify signal change across force. In (B), the first and second rows show SNR under 2 N and 10 N loads, respectively. The left and right columns represent tests with varying amplitude (constant period) and period (constant amplitude). The diamond markers connected by a solid line represent the mean SNR for each material.}
    \label{fig:sensitivity}
\end{figure}

\subsubsection{Force Sensitivity}

SI provides the highest force sensitivity at low loads ($<10$ N), as indicated by the slope magnitudes in Fig. \ref{fig:sensitivity}A, but is fully saturated beyond this point. PU30 shows force sensitivity similar to that of SI and maintains a clean signal up to 20 N but exhibits a large loading/unloading hysteresis band. Anecdotally, we noted that Clear Flex 30 (for PU30) exhibits much greater viscoelasticity than Solaris (SI) and Clear Flex 50 (PU50), which supports these results. The PU50 gel provides a lower but more consistent sensitivity across the entire loading range, indicating its usefulness for higher load applications.

\subsubsection{Spatial Sensitivity} 

The results from our spatial sensitivity test, shown in Fig. \ref{fig:sensitivity}B, illustrate the performance of the materials across the different ridged surfaces and loading conditions. With a 2 N load, SI generally matches or outperforms PU. For constant period tests, all gels demonstrate a gradual increase in performance with increasing amplitude, with SNR plateauing at 10-15 dB for amplitudes larger than 0.03 mm. For constant amplitude tests, SI provides much clearer signals, especially for periods of 1.0 mm or smaller. When the load is increased to 10 N, the SNR increases for PU30 and PU50 but reduces for silicone, resulting in better relative performance for polyurethane across most ridges. The reduction of SI performance at higher load can be attributed to an increase in noise power and may not reflect a decrease in the sensor's ability to detect the ridges, discussed further in Section \ref{sec:discussion}. Both materials show the same general trends across surfaces, seeing reductions in performance at amplitudes and periods less than 0.035 mm and 1.1 mm, respectively. Interestingly, PU50 provides a signal similar to or greater than PU30 across the majority of tests. Similar to SI, we suspect that the lower stiffness of PU30 leads to a stronger noise floor, which reduces the SNR.

\section{Discussion}
\label{sec:discussion}

Our results show that the polyurethane gels, with higher hardness and better adhesion to acrylic, greatly outlast silicone-based gels across multiple types of wear. Additionally, polyurethane extends the effective force-sensing range in comparison to silicone. We propose that material selection should be application-driven: silicone remains ideal for low-force tasks requiring high precision; PU30 offers improved durability for a similar level of precision (though its hysteresis should be considered); and PU50 excels for higher-force, rugged environments where reliability outweighs sub-millimeter sensitivity.

We select the silicone gel formulation due to its representativeness of common VBTS solutions (e.g. Digit, GelSight) and ready availability, and the polyurethane formulations based on the softest available water-clear polyurethane rubbers. Higher hardness silicones are available from industrial vendors with long lead times and higher prices and are therefore omitted from this study. However, to fully understand the individual benefits of material and hardness changes, future work should perform these tests across a wider range of gel compositions.

We note that our evaluation techniques face certain limitations. Our resilience tests--while designed to be representative of real wear--are also highly controlled, with specific contact geometries and loading forces that are not comprehensive. Thus, we do not expect the test lifetimes of our sensors to directly represent their cyclic lifetime in application, and we primarily emphasize the relative performance of each material in these tests. The $MAE$ used in our resilience tests measures raw sensor change without characterizing how this change affects sensor performance. It is possible that performance can be recovered through sensor recalibration or better models, but we leave this to future work. 

Finally, our learning-free spatial sensitivity metric is limited by the effect of bulk deformation at high loads. Larger forces increase the bulk deformation of the sensor, generating large changes in the sensor image that remain after background subtraction. These changes can increase our noise power measurement across frequency bins, leading to reduced SNR for softer materials at high load. Future work will explore approaches to reduce this observed effect.

\section{Real-World Translation}
\label{sec:demo}

\begin{figure}
    \centering
    \vspace{3mm}
    \includegraphics[width=85mm]{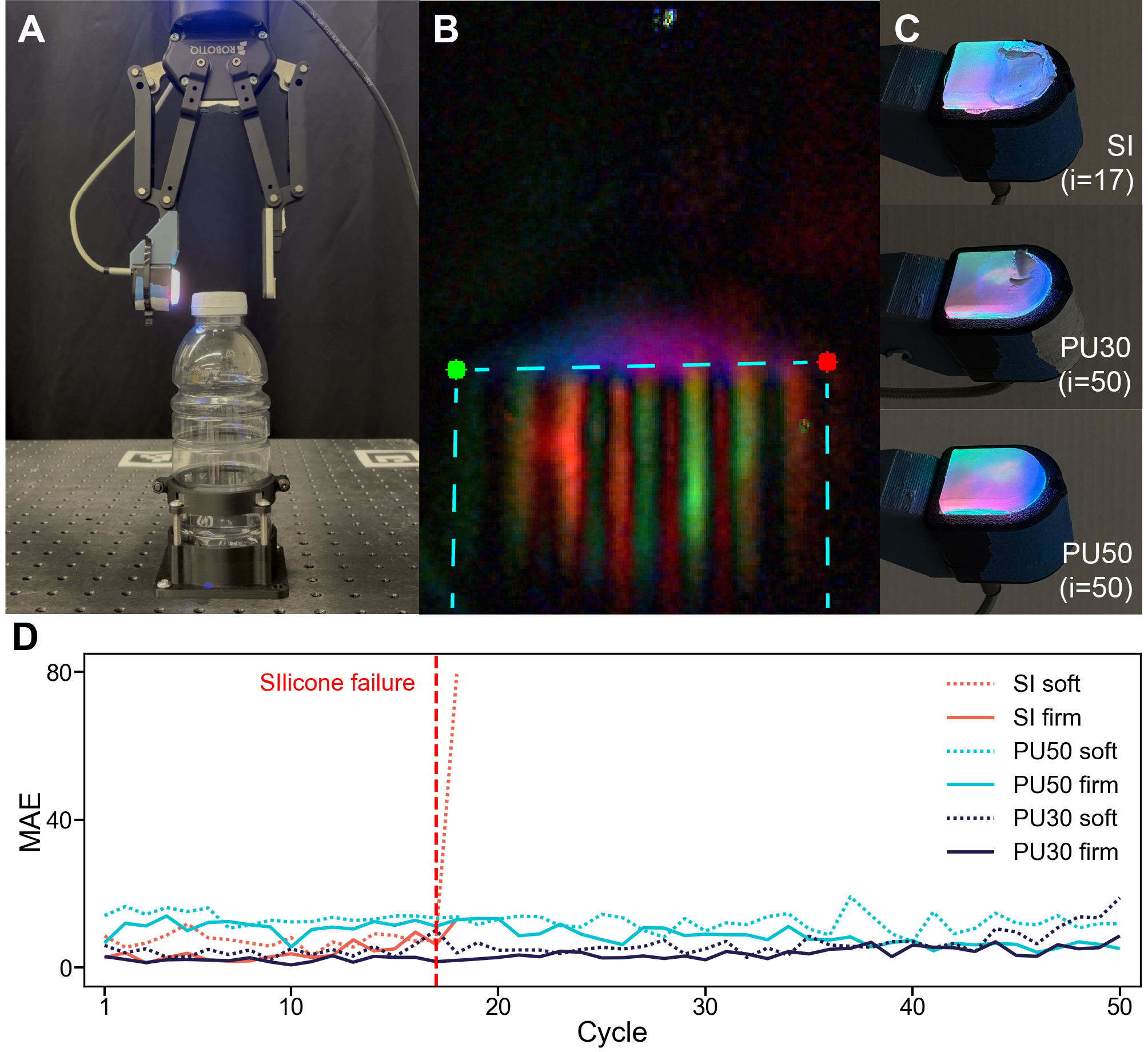}
    \caption{(A) For demonstration, we perform repeated loosening and tightening of a bottle cap with a sensorized gripper. (B) A trained CNN predicts two points that define a simplified contact region shown with dashed lines. (C) Gels exhibit varying degrees of damage after completing $i$ cycles. (D) MAE of model predictions are calculated for the soft and firm grasps performed during tightening at each cycle.}
    \label{fig:demo}
\end{figure}

To ground our controlled framework in a real task, we perform a bottle cap loosening and tightening task that induces a combination of compression, shear, and abrasion (Fig. \ref{fig:demo}A). We use a Robotiq 2F-140 parallel jaw gripper with one Digit sensor attached to a UR-10 robot arm. To measure sensor performance, a lightweight convolutional network trained on 243 contact images predicts the coordinates of two hand-annotated points defining a simplified contact region (Fig. \ref{fig:demo}B). 

We conduct up to fifty cycles of the task, tightening to a max torque of 0.8 N-m. For variability, retightening is initiated with a soft grasp and finished with a firm grasp. The trained model predicts the resulting grasp contact regions, which are compared to hand annotations.

As shown by the mean absolute error (MAE) of the model predictions, the silicone gel fails via tearing (Fig. \ref{fig:demo}C,D) after the 17th cycle of tightening and exhibits an immediate reduction in performance. After failure, the gel does not provide enough friction to twist the cap, resulting in an early stoppage. For the PU30 gel, a tear occurs in the image transfer layer halfway through the test and continues to grow, only impacting performance for soft grasps after 40 cycles. PU50 successfully completes all 50 cycles, maintaining consistent performance without visible failure--although with slightly worse performance than the other two materials. This demonstration confirms that the material trade-offs quantified in our framework relate directly to hardware resilience and signal sensitivity in an applied manipulation task.

\section{Conclusion}
\label{sec:conclusion}

The practical deployment of vision-based tactile sensors in real-world environments is often hindered by a lack of physical sensor resilience, and efficient hardware innovation is bottlenecked by a lack of evaluations for intrinsic sensor performance. In this work, we introduce polyurethane gels as more resilient alternatives and propose a methodological framework containing repeatable, learning-free evaluations to compare their resilience and sensitivity to a representative silicone baseline.

The results of our resilience evaluations show that polyurethane gels can indeed outlast silicone gels across repeated cycles of compression, shear, and abrasion, delaying or resisting the catastrophic punctures and tears experienced by many of the silicone samples. While our learning-free sensitivity analysis confirms that silicone provides enhanced force and spatial sensitivity compared to polyurethane at low load, it also reveals that polyurethane can provide high spatial sensitivity and better force sensitivity with increased load.

The use of silicone or polyurethane gels should depend on the application. For delicate manipulation tasks where high sensitivity to forces and contact geometries are the priority, silicone remains a suitable option. However, for applications where sensors must be deployed into different unstructured environments without easy access to replacements, we suggest that polyurethane gels offer an advantage that scales with hardness. However, our study is limited to three gel material configurations. Further exploration of materials and fabrication techniques may yield better solutions, and we propose that our resilience and sensitivity evaluation techniques provide simple, consistent ways to compare across them.

\section*{Acknowledgment}

The authors would like to thank Roberto Calandra for his guidance throughout the project, and Gia Jeong for her help with the demonstration. The authors also acknowledge the support of the members of the Embodied Dexterity Group.

%%%%%%%%%%%%%%%%%%%%%%%%%%%%%%%%%%%%%%%%%%%%%%%%%%%%%%%%%%%%%%%%%%%%%%%%%%%%%%%%

\bibliographystyle{ieeetr}
\bibliography{ref}

\end{document}